# Face Recognition by Fusion of Local and Global Matching Scores using DS Theory: An Evaluation with Uni-classifier and Multi-classifier Paradigm

Dakshina R. Kisku
Dr. B. C. Roy Engg.
College, India
`drkisku@ieee.org`

Massimo Tistarelli
University of Sassari,
Alghero (SS), Italy
`tista@uniss.it`

Jamuna Kanta Sing
Jadavpur University,
Kolkata, India
`jksing@ieee.org`

Phalguni Gupta
I.I.T Kanpur, Kanpur,
India
`pg@cse.iitk.ac.in`

## Abstract

*Faces are highly deformable objects which may easily change their appearance over time. Not all face areas are subject to the same variability. Therefore decoupling the information from independent areas of the face is of paramount importance to improve the robustness of any face recognition technique. This paper presents a robust face recognition technique based on the extraction and matching of SIFT features related to independent face areas. Both a global and local (as recognition from parts) matching strategy is proposed. The local strategy is based on matching individual salient facial SIFT features as connected to facial landmarks such as the eyes and the mouth. As for the global matching strategy, all SIFT features are combined together to form a single feature. In order to reduce the identification errors, the Dempster-Shafer decision theory is applied to fuse the two matching techniques. The proposed algorithms are evaluated with the ORL and the IITK face databases. The experimental results demonstrate the effectiveness and potential of the proposed face recognition techniques also in the case of partially occluded faces or with missing information.*

## 1. Introduction

Face recognition is one of most challenging research areas in machine vision and biometrics [1, 2]. The variability in the appearance of face images, either due to intrinsic and extrinsic factors, makes the identification problem ill-posed and difficult to solve. Moreover, additional complexities like the data dimensionality and the motion of face parts causes major changes in appearance. In order to make the problem well-posed, vision researchers have adapted and applied an abundance of algorithms for pattern classification, recognition and learning. To cope for the data dimensionality, several appearance-based techniques have been successfully used, such as the Principal Component Analysis (PCA) [1], Linear Discriminant Analysis (LDA) [1], Fisher Discriminant Analysis (FDA) [1], and Independent Component Analysis (ICA) [1]. Other methods have been studied based on the extraction of salient facial features by means of cascaded scale-space filtering [3, 4, 5, 6]. Most of the times, one missing part is the link between the features extracted from the face images and the geometry of the face itself.

The aim of this paper is to perform a robust and cost effective face recognition using SIFT features extracted from face images [7, 8, 9, 10] but also directly related to the face geometry. In this regard, two face-matching techniques, based on local and global information and their fusion are proposed. In the local matching strategy, SIFT keypoint features are extracted from face images in the areas corresponding to facial landmarks, such as the eyes, nose and mouth. Facial landmarks are automatically located by means of a standard facial landmark detection algorithm [11, 12]. Then matching of a pair of feature vectors is performed by a minimum Euclidean distance metric. Matching scores produced from each pair of salient features are fused together using the sum rule [13]. In the global matching strategy, the SIFT features extracted from the facial landmarks are fused together by concatenation. Also in this case, matching is performed by means of a minimum Euclidean distance metric. The matching scores obtained from the local and global strategies are fused together using the Dempster-Shafer decision theory. The proposed techniques are evaluated with two face databases, the IITK and ORL (formerly known as AT&T) face databases.

The paper is organized as follows. Section 2 briefly describes the SIFT features extraction. Local and global matching strategies are discussed in Section 3. Section 4 describes the fusion of local and global matching using the Dempster-Shafer theory. The experimental results are presented and discussed in Section 5 and 6.

## 2. Overview of the SIFT feature extraction

The scale invariant feature transform, called SIFT descriptor, has been proposed by Lowe [8, 9] and proved to be invariant to image rotation, scaling, translation, partly illumination changes. The basic idea of the SIFT descriptor is detecting feature points efficiently through a staged filtering approach that identifies stable points in the scale-space. Local feature points are extracted by searching peaks in the scale-space from a difference of Gaussian (DoG) function. The feature points are localized using the measurement of their stability and orientations are assigned based on local image properties. Finally, the feature descriptors, which represent local shape distortions and illumination changes, are determined.

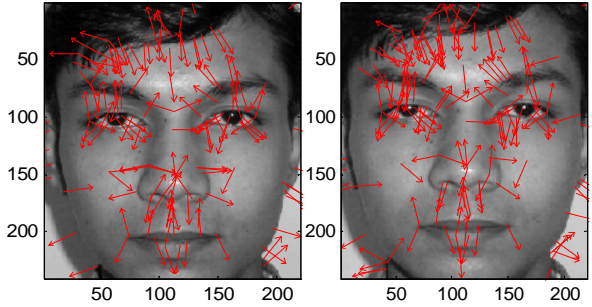

Figure 1: Invariant SIFT feature extraction are shown on a pair of face images.

Each feature point is composed of four types of information – spatial location ($x,y$), scale ($S$), orientation ($\theta$) and Keypoint descriptor ($K$). For the sake of the experimental evaluation, only the keypoint descriptor [6, 9, 10] has been taken into account. This descriptor consists of a vector of 128 elements representing the orientations within a local neighborhood. In Figure 1, the SIFT features extracted from a pair of face images are shown.

## 3. Local and global matching

In this section we develop two matching strategies, namely the local, based on parts, and the global face matching. In addition, we introduce a classifier fusion technique, where the scores obtained from the local strategy are fused together in terms of matching scores obtained from individual classifiers.

### 3.1. Local face matching strategy

Faces are deformable objects which are generally difficult to characterize with a rigid representation. Different facial regions, not only convey different information on the subject's identity, but also suffer from different time variability either due to motion or illumination changes [14]. A typical example is the case of a talking face. While the eyes can be almost still and invariant over time, the mouth moves changing its appearance over time. As a consequence, the features extracted from the mouth area cannot be directly matched with the corresponding features from a static template. Moreover, single facial features may be occluded making the corresponding image area not usable for identification. For these reasons to improve the robustness of the identification process it is mandatory to decouple the image information corresponding to different face areas. The aim of the proposed local matching technique is to correlate the extracted SIFT features with independent facial landmarks. The SIFT descriptors are extracted and grouped together at locations corresponding to static (eyes, nose) and dynamic (mouth) facial positions.

In Figure 2 and 3 an example showing the concept of independent matching facial features from local areas is presented.

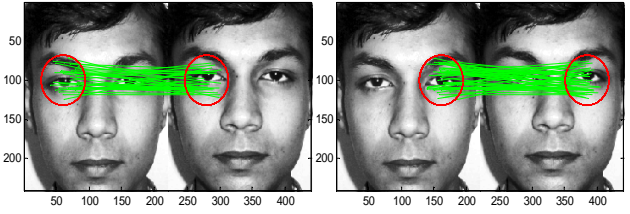

Figure 2: Example of matching static facial features.

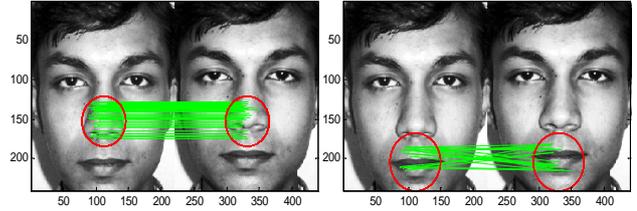

Figure 3: Example of independent matching of static and dynamic facial features.

The eyes and mouth positions are automatically located by applying the technique proposed in [11]. The position of nostrils is automatically located by applying the technique proposed in [12]. A circular region of interest (ROI), centered at each extracted facial landmark location, is defined to determine the SIFT features to be considered as belonging to each face area.

Given a face image $I$ four independent ROI are extracted. Two ROI regions $I^{left\text{-}eye}$ and $I^{right\text{-}eye}$ refer to the left and right eyes. Two other ROI regions, $I^{nose}$ and $I^{mouth}$ refer to the nose and mouth locations. The SIFT feature points are then extracted from these four regions and gathered together into four groups. From these groups pair-wise salient feature matching is performed. Finally, the matching scores obtained are fused together by the sum fusion rule [13] and the fused score are compared against a threshold. More formally, if $D^{Left-eye}(I^{gallery}, I^{gallery})$ is the distance between a pair of left eyes, then the distance can be defined as follows:

$$D^{Left-eye}(I^{test}, I^{gallery}) = \sqrt{\sum_{i \in m, j \in n}(I_j^{test}(k_{left-eye}) - I_i^{gallery}(k_{left-eye}))^2} \leq \Psi_{left-eye}^k \quad (1)$$

where, $m$ and $n$ are the dimensions of concatenated feature points for a pair of gallery and test samples and $k$ refers the keypoint descriptor. $\Psi^k$ is the threshold, which is computed a priori from a training set of face images. This face set is disjoint from the image sets used for testing and validation.

In the same fashion, the distances for a pair of right eyes, for a pair of noses and for a pair of mouths can be determined as follows:

$$D^{Right-eye}(I^{test}, I^{gallery}) = \sqrt{\sum_{i \in m, j \in n}(I_j^{test}(k_{right-eye}) - I_i^{gallery}(k_{right-eye}))^2} \quad (2)$$

$$\leq \Psi_{right-eye}^k$$

$$D^{nose}(I^{test}, I^{gallery}) = \sqrt{\sum_{i \in m, j \in n}(I_j^{test}(k_{nose}) - I_i^{gallery}(k_{nose}))^2} \quad (3)$$

$$\leq \Psi_{nose}^k$$

$$D^{mouth}(I^{test}, I^{gallery}) = \sqrt{\sum_{i \in m, j \in n}(I_j^{test}(k_{mouth}) - I_i^{gallery}(k_{mouth}))^2} \quad (4)$$

$$\leq \Psi_{mouth}^k$$

Finally, the fused matching score $FD(I^{test}, I^{gallery})$ is computed by combining these four individual matching scores together using sum rule [13]:

$$FD_{LOCAL}(I^{test}, I^{gallery}) = sum(D^{Left-eye}(I^{test}, I^{gallery}), \ldots \quad (5)$$
$$D^{Right-eye}(I^{test}, I^{gallery}), D^{nose}(I^{test}, I^{gallery}), D^{mouth}(I^{test}, I^{gallery}))$$

### 3.2. Global face matching strategy

While for the local matching each face area is handled independently, in the global matching all SIFT features are grouped together. In particular, the SIFT features extracted from the image areas corresponding to the located four facial landmarks, are grouped together to form an augmented vector by concatenation. The actual matching is performed comparing the global feature vectors for a pair of face images. Before performing the face matching a one to one correspondence is established for each pair of facial landmarks, as discussed in Section 3.1.

In order to compute the matching distance/score between gallery and probe samples by computing distance between a pair of concatenated feature sets, let us consider, $I^{Left-eye}(k), I^{Right-eye}(k), I^{nose}(k), I^{mouth}(k)$ which are the four facial features computed from both the gallery and probe face images. Two concatenated keypoint sets can be computed as:

$$I_{gallery}^{Left-eye}(k_i) = \{I_{gallery}^{Left-eye}(k_1), I_{gallery}^{Left-eye}(k_2), \ldots, I_{gallery}^{Left-eye}(k_m)\} \quad \because i \in m;$$

$$I_{gallery}^{Right-eye}(k_i) = \{I_{gallery}^{Right-eye}(k_1), I_{gallery}^{Right-eye}(k_2), \ldots, I_{gallery}^{Right-eye}(k_n)\} \quad \because i \in n;$$

$$I_{gallery}^{nose}(k_i) = \{I_{gallery}^{nose}(k_1), I_{gallery}^{nose}(k_2), \ldots, I_{gallery}^{nose}(k_p)\} \quad \because i \in p;$$

$$I_{gallery}^{mouth}(k_i) = \{I_{gallery}^{mouth}(k_1), I_{gallery}^{mouth}(k_2), \ldots, I_{gallery}^{mouth}(k_q)\} \quad \because i \in q;$$

where, $m$, $n$, $p$ and $q$ are the dimensions of the extracted keypoint feature sets computed from the left eye, right eye, nose and mouth. In order to obtain a fused feature set for a gallery sample face, we concatenate the keypoints of four components together, one by one, as:

$$I^{gallery}(k) = \{I_{gallery}^{Left-eye}(k_m) \cup I_{gallery}^{Right-eye}(k_n) \cup \ldots \quad (6)$$
$$I_{gallery}^{nose}(k_p) \cup I_{gallery}^{mouth}(k_q)\};$$

Similarly, the concatenated feature set for a probe sample is obtained by the equation:

$$I^{probe}(k) = \{I_{probe}^{Left-eye}(k_{m'}) \cup I_{probe}^{Right-eye}(k_{n'}) \cup \ldots \quad (7)$$
$$I_{probe}^{nose}(k_{p'}) \cup I_{probe}^{mouth}(k_{q'})\};$$

The final matching score $FD_{GLOBAL}(I^{probe}, I^{gallery})$ is computed by first determining all the minimum pair distances and then computing a mean score of all the minimum pair distances as:

$$FD_{GLOBAL}(I^{probe}, I^{gallery}) = \sqrt{\sum_{i \in M} \min_{j \in N}\{\min\{I^{gallery}(k_i), I^{probe}(k_j)\}\}} \quad (8)$$

In Equation (8), the final distance is determined by the Hausdorff distance metric and the distance score is compared against a threshold computed heuristically from a training set of face images. As for the local matching threshold, this face set is disjoint from the image sets used for testing and validation.

## 4. Fusion of local and global matching scores using the Dempster-Shafer Theory

In the proposed classifier fusion, the Dempster-Shafer decision theory [15, 16, 17] is applied to combine the decision of the local and global matching.

The Dempster-Shafer theory is based on combining the evidences obtained from different sources to compute the probability of an event. This is obtained combining three elements: the basic probability assignment function (*bpa*), the belief function (*bf*) and the plausibility function (*pf*).

The *bpa* maps the power set to the interval [0,1]. The bpa function of the empty set is 0 and the *bpa's* of all the subsets of the power set is 1. Let m denote the *bpa* function and *m(A)* represent the *bpa* for a particular set *A*. An element of a universal set *X* belongs to the set *A*, but to no particular subset of *A*, while *m(A)* would represent the proportion of all the relevant evidence and claims the association of the element to the set *A*. The value of *m(A)* pertains only to the set *A* and makes no association to any subsets of *A*. If we consider *m(B)* is the *bpa* for another set *B* and $B \subset A$, then we can say that any further evidence happens to the subsets of *A*. Formally, the basic probability assignment function can be represented by the following equations

$$m : P(X) \to [0,1] \quad (9)$$
$$m(\phi) = 0 \quad (10)$$
$$\sum_{A \in P(X)} m(A) = 1 \quad (11)$$

Where $P(X)$ is the power set of *A*, Ø is the empty set, and *A* is a set in the power set $A \in P(X)$.

From the basic probability assignment, the upper and lower bounds of an interval are bounded by two nonadditive continuous measures called Belief and Plausibility. The lower bound Belief for a set *A* is defined as the sum of all the basic probability assignments of the proper subsets *(B)* of the set of interest *(A)* ($B \subset A$). The upper bound Plausibility is the sum of all the basic probability assignments of the sets *(B)* that intersect *(A)* $(B \cap A) \neq \emptyset$. For all the sets *A* that are elements of the power set $(A \in P(X))$:

$$Bel(A) = \sum_{B \mid B \subseteq A} m(B) \quad (12)$$
$$Pl(A) = \sum_{B \mid B \cap A \neq \emptyset} m(B) \quad (13)$$

An inverse function with the Belief measures can be used to obtain the basic probability assignment:

$$m(A) = \sum_{B \mid B \subseteq A} (-1)^{\gamma} Bel(B) \quad \because \gamma = |A - B| \quad (14)$$

where *|A-B|* is the difference of the cardinality between the two sets *A* and *B*.

It is possible to derive these two measures, Belief and Plausibility from each other. If Plausibility can be derived from Belief measures, then the following equation holds:

$$Pl(A) = 1 - Bel(\overline{A}) \quad (15)$$

Where $\overline{A}$ is the complement of *A*. In addition, the Belief measures can be written as:

$$Bel(\overline{A}) = \sum_{B \mid B \subseteq A} m(B) = \sum_{B \mid B \cap A = \emptyset} m(B) \quad (16)$$

and

$$\sum_{B \mid B \cap A \neq \emptyset} m(B) = 1 - \sum_{B \mid B \cap A = \emptyset} m(B) = Pl(A) \quad (17)$$

Let, $\Gamma^{Local}$ and $\Gamma^{Global}$ are the two matching score sets obtained from the two different matching processes. Also, let, $m(\Gamma^{Local})$ and $m(\Gamma^{Global})$ are the bpa functions for the Belief measures $Bel(\Gamma^{Local})$ and $Bel(\Gamma^{Global})$ for the two classifiers, respectively. Then the Belief probability assignments *(bpa)* $m(\Gamma^{Local})$ and $m(\Gamma^{Global})$ can be combined together to obtained a Belief committed to a matching score set $C \in \Theta$ according to the following combination rule or orthogonal sum rule

$$m(C) = m(\Gamma^{Local}) \oplus m(\Gamma^{Global}) = \frac{\sum_{\Gamma^{Local} \cap \Gamma^{Global} = C} m(\Gamma^{Local}) m(\Gamma^{Global})}{1 - \sum_{\Gamma^{Local} \cap \Gamma^{Global} = \emptyset} m(\Gamma^{Local}) m(\Gamma^{Global})}, \quad C \neq \emptyset. \quad (18)$$

The denominator in equation (18) is a normalizing factor which denotes how much the Belief probability assignments $m(\Gamma^{Local})$ and $m(\Gamma^{Global})$ are conflicting.

Let *m(Local)* and *m(Global)* are the two matching score sets obtained from the local and global matching strategies. They can be fused together recursively as:

$$m(final) = m(Local) \oplus m(Global) \quad (19)$$

where, $\oplus$ denotes the Dempster combination rule. The final decision of user acceptance and rejection can be established by the following equation and by applying the threshold $\Psi$ to *m(final)*

$$decision = \begin{cases} accept, & if \quad m(final) \geq \Psi \\ reject, & otherwise \end{cases} \quad (20)$$

## 5. Experimental evaluation and results

To investigate the effectiveness of the proposed local and global face matching strategies and their fusion, we carried out extensive experiments on the IITK and the ORL face databases [18].

The local and global matching strategies are evaluated independently on both databases. The matching scores obtained from the proposed techniques (local and global matching) are fused together to improve the recognition performance.

### 5.1. Evaluation on the IITK database

The IITK face database consists of 800 face images with four images per subject (200X4). The images have been captured in different sessions, within a controlled environment with ±20 degrees of rotation in the head pose and with almost uniform lighting. The facial expressions are consistently kept neutral, with very small changes. The face images are downscaled to 140×100 pixels. For the face matching, all probe images are matched against all target images, yielding 800×3 genuine scores (images from the same subject) and 800×799×3 imposter scores (images from different subjects).

with IITK database. In total 200×4 genuine scores and 200×4×199 imposter scores were generated for the entire data set.

Also in this case, the results obtained from the local matching strategy outperform the global matching strategy in terms of recognition rates and true/false acceptance.

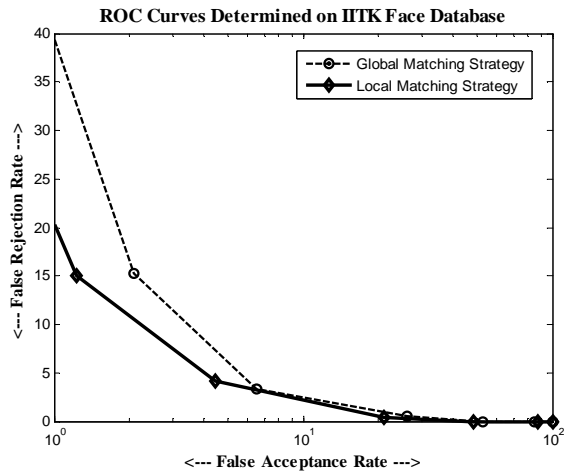

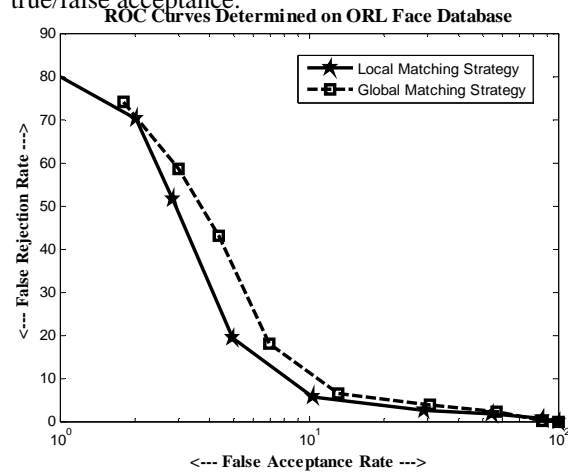

Figure 4: ROC curves determined on IITK face database is shown for both the local and global matching strategy.

| Matching Strategy | FRR (%) | FAR (%) | EER (%) | Recognition rate (%) |
|---|---|---|---|---|
| Local matching | 6.29 | 2.19 | 4.24 | 95.76 |
| Global matching | 9.87 | 3.61 | 6.79 | 93.21 |

Table 1. Performance metrics computed from the tested matching strategies as obtained from the IITK database.

The results obtained from the IITK dataset are quite promising. From the ROC curve in Figure 4 and the Table 1, it turns out that the fusion of pair-wise local matching of the facial feature components outperforms the global matching strategy. This result clearly shows the advantages of component-based strategies to cope for unexpected changes in few areas of the face. The fusion of local information allows to achieve a robust identification.

### 5.2. Evaluation on the ORL database

The same recognition experiment as before, was performed on the ORL face database (formerly known as AT&T face database) [18]. The ORL face database consists of 400 images taken from 40 persons. Out of these 400 images, we used 200 face images (5 samples per subject) in which ±20 to ±30 degrees orientation changes have been considered. The face images show variations of pose and facial expression (smile/not smile, open/closed eyes). When the faces were taken, the original resolution was 92 x 112 pixels. However, for our experiment the resolution was re-scaled to 140×100 pixels in line

Figure 5: ROC curves determined on ORL face database are shown for local and global matching strategies.

| Matching Strategy | FRR (%) | FAR (%) | EER (%) | Recognition rate (%) |
|---|---|---|---|---|
| Local matching | 3.77 | 1.45 | 2.61 | 97.39 |
| Global matching | 5.86 | 2.48 | 4.17 | 95.83 |

Table 2. Performance metrics computed from the tested matching strategies as obtained from the ORL database.

### 5.3. Fusion of global and local matching scores

In order to determine the effectiveness of score level fusion of local and global face matching, we applied the Dempster-Shafer theory for fusion. Before performing score level fusion, the computed matching scores are firstly normalized by applying the "min-max" technique [13]. The Dempster-Shafer decision theory is then applied to the normalized.

The fusion method has been applied to the IITK and the ORL face databases. To limit the page length, the partial results for Yale database have not been included instead the DS theory based fusion result. Yet, the large variability in the face samples due to changes in the facial expression allowed us to thoroughly validate the advantage of a part-based representation and matching.

The ROC curves of the error rates obtained from the score fusion applied to the three face databases are shown in figure 6. The computed recognition accuracy for the IITK database is 96.29% and for the

ORL database is 98.93%. These values correspond to an improvement in accuracy of about 1% over the local fusion strategy. The recognition result for the Yale database is 98.19%, which is also promising.

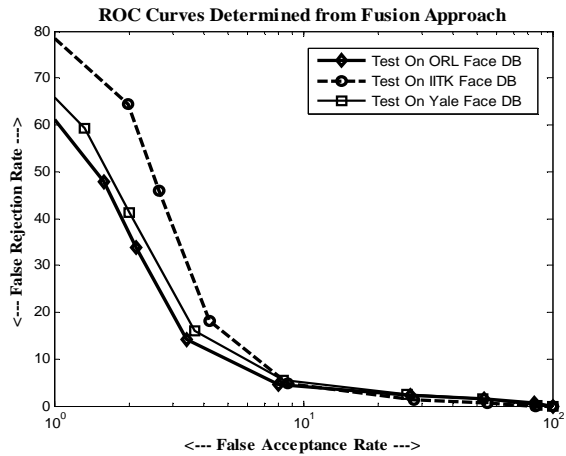

Figure 6: ROC curves determined from three face databases: IITK, ORL and Yale face databases.

## 6. Conclusion

Human faces can be characterized both on the basis of local as well as of global features. While global features are easier to capture they are generally less discriminative than localized features, but are less sensitive to localized changes in the face due to the partial deformability of the facial structure. On the other hand, local features on the face can be highly discriminative, but may suffer for local changes in the facial appearance or partial face occlusion. The optimal face representation should then allow to match localized facial features, but also determining a global similarity measurement for the face [14].

In this paper, a robust, integrated classification paradigm for face recognition has been presented, comparing a local and a global face representation. Both representations are based on robust and invariant photometric features (SIFT features). Performances of individual matching techniques have been evaluated with two face databases: the IITK and the ORL face datasets. Results on score level fusion are reported also from the Yale database.

The Dempster-Shafer theory has been applied to fuse both the local and global fusion strategies. The experiments performed on the ORL face database scored 98.93% of accuracy in recognition. The same tests performed on the IITK face dataset scored 96.29% of accuracy in recognition of about 3% increase in relative performances over the global matching method alone.